\PassOptionsToPackage{table,dvipsnames}{xcolor}
\PassOptionsToPackage{sort&compress}{natbib}
\documentclass{article}

\usepackage{opsdv_preprint,times}
\usepackage[letterpaper,left=0.9in,right=0.9in,top=0.68in,bottom=0.72in]{geometry}
\usepackage{titlesec}

\usepackage{amsmath,amsfonts,bm}

\def\eqref#1{equation~\ref{#1}}
\def\1{\bm{1}}

\DeclareMathAlphabet{\mathsfit}{\encodingdefault}{\sfdefault}{m}{sl}
\SetMathAlphabet{\mathsfit}{bold}{\encodingdefault}{\sfdefault}{bx}{n}

\usepackage{iftex}
\ifPDFTeX
\usepackage[utf8]{inputenc}
\fi
\usepackage[T1]{fontenc}
\usepackage{amsmath}
\usepackage{amssymb}
\usepackage{booktabs}
\usepackage{enumitem}
\usepackage{graphicx}
\usepackage{hyperref}
\usepackage{url}
\usepackage[capitalize,nameinlink]{cleveref}
\usepackage{multirow}
\usepackage{xcolor}
\usepackage{xspace}
\usepackage{tcolorbox}
\usepackage{caption}
\usepackage{placeins}
\usepackage{float}

\hypersetup{
    colorlinks=true,
    linkcolor=blue,
    urlcolor=blue,
    citecolor=blue,
    pdftitle={MSEditor: Toward Consistent Multi-Shot Video Editing},
    pdfauthor={Kunyu Feng et al.}
}

\definecolor{Blue1}{HTML}{003366}
\definecolor{Blue2}{HTML}{004080}
\definecolor{Blue3}{HTML}{0059B3}
\definecolor{Blue4}{HTML}{0073E6}
\definecolor{Blue5}{HTML}{3399FF}
\definecolor{Blue6}{HTML}{66B3FF}
\definecolor{Blue7}{HTML}{0099CC}
\definecolor{Blue8}{HTML}{00CCCC}
\definecolor{frontgray}{HTML}{F1F2F3}
\definecolor{frontdark}{HTML}{4B4B4B}

\newcommand{\blueMSeditor}{%
    \textcolor{Blue1}{M}%
    \textcolor{Blue2}{S}%
    \textcolor{Blue3}{E}%
    \textcolor{Blue4}{d}%
    \textcolor{Blue5}{i}%
    \textcolor{Blue6}{t}%
    \textcolor{Blue7}{o}%
    \textcolor{Blue8}{r}%
}

\newtcolorbox{frontmatterbox}{
    colback=frontgray,
    colframe=frontgray,
    boxrule=0pt,
    arc=8pt,
    left=20pt,
    right=20pt,
    top=18pt,
    bottom=18pt,
    width=\textwidth,
    before skip=0pt,
    after skip=0pt
}

\titleformat{\section}{\sffamily\Large\bfseries}{\thesection}{1em}{}
\titleformat{\subsection}{\sffamily\large\bfseries}{\thesubsection}{0.8em}{}
\titleformat{\subsubsection}{\sffamily\normalsize\bfseries}{\thesubsubsection}{0.8em}{}
\titlespacing*{\section}{0pt}{2.0ex plus 0.4ex minus 0.2ex}{1.1ex}
\titlespacing*{\subsection}{0pt}{1.7ex plus 0.3ex minus 0.2ex}{0.8ex}
\titlespacing*{\subsubsection}{0pt}{1.4ex plus 0.3ex minus 0.2ex}{0.6ex}

\newcommand{\papersubtitle}{Toward Consistent Multi-Shot Video Editing}
\newcommand{\paperfrontsubtitle}{Toward Consistent Multi-Shot Video Editing}
\title{\blueMSeditor: \papersubtitle}

\author{
Kunyu Feng\textsuperscript{1,*},
Yue Ma\textsuperscript{2,*},
Bingyuan Wang\textsuperscript{1},
Yuefeng Wang\textsuperscript{3},
Zhiyuan Qin\textsuperscript{4},
Hao Cheng\textsuperscript{5},
Hao Li\textsuperscript{4},
Qifeng Chen\textsuperscript{2},
Zeyu Wang\textsuperscript{1,2}
}

\newcommand{\frontauthorblock}{%
\textbf{Kunyu Feng\textsuperscript{1,*},
Yue Ma\textsuperscript{2,*},
Bingyuan Wang\textsuperscript{1},
Yuefeng Wang\textsuperscript{3},
Zhiyuan Qin\textsuperscript{4}}\\
\textbf{Hao Cheng\textsuperscript{5},
Hao Li\textsuperscript{4},
Qifeng Chen\textsuperscript{2},
Zeyu Wang\textsuperscript{1,2}}\\[0.08in]
\textsuperscript{1}HKUST (Guangzhou) \quad
\textsuperscript{2}HKUST \quad
\textsuperscript{3}Baidu Inc.\\
\textsuperscript{4}Beijing Innovation Center of Humanoid Robotics \quad
\textsuperscript{5}Tsinghua University\\
\textsuperscript{*}Equal contribution.%
}

\newcommand{\paperabstract}{%
In this paper, we tackle the problem of performing consistent, unified modifications to a multi-shot video sequence. This task is particularly challenging because multi-shot videos consist of discontinuous temporal segments that vary significantly in viewpoint, camera scale, and subject pose, leading to severe identity drift and cumulative error propagation. Achieving coherent edits requires establishing reliable cross-shot semantic awareness to maintain stable subject appearance and visual continuity across these disjointed boundaries. To address this, we propose MSEditor, the first framework designed specifically for consistent multi-shot video editing. To overcome the scarcity of high-quality multi-shot training data, we repurpose existing multi-view video datasets to provide robust cross-shot supervision. Architecturally, we introduce a Supervisory Adapter that injects this cross-shot information into the diffusion backbone, enabling the model to learn identity-consistent representations. Furthermore, to effectively mitigate cumulative errors and ensure long-range temporal coherence, we design a Cross-Shot Packing strategy that dynamically aggregates information from semantically related shots within the self-attention window. Extensive experiments demonstrate that MSEditor significantly outperforms existing methods on our curated multi-shot video editing benchmark in terms of identity preservation, temporal stability, and overall visual quality.
}

\newcommand{\makefrontmatter}{%
\vspace*{0.16in}
\begin{frontmatterbox}
\begin{minipage}[c]{0.68\linewidth}
{\sffamily\bfseries\fontsize{28}{31}\selectfont \blueMSeditor\par}
\vspace{0.08in}
{\sffamily\bfseries\Large\color{frontdark}\paperfrontsubtitle\par}
\end{minipage}%
\hfill
\begin{minipage}[c]{0.26\linewidth}
\centering
\includegraphics[height=0.78in,trim=103 49 112 45,clip]{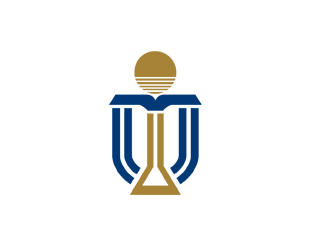}%
\hspace{0.10in}%
\includegraphics[height=0.78in]{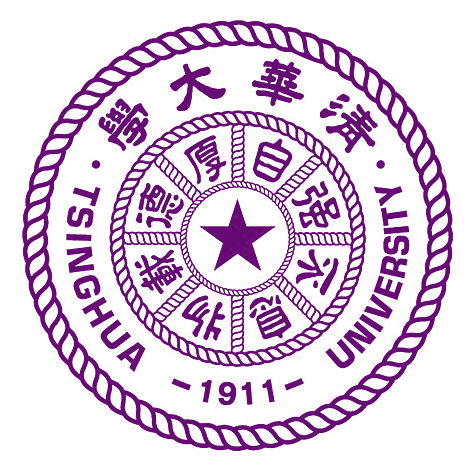}
\end{minipage}
\par\vspace{0.22in}
{\normalsize\frontauthorblock\par}
\vspace{0.26in}
{\normalsize\paperabstract\par}
\end{frontmatterbox}
\vspace{0.22in}
}

\setcitestyle{numbers,square,sort&compress,comma}
\let\cite\citep

\arxivcopy
\renewcommand{\headrulewidth}{0pt}
\renewcommand{\footrulewidth}{0pt}
\AddToShipoutPicture{%
  \AtPageLowerLeft{%
    \raisebox{0.34in}{\makebox[\paperwidth][c]{\thepage}}%
  }%
}

\begin{document}

\makefrontmatter

\begin{figure}[t]
  \centering
  \includegraphics[width=\textwidth]{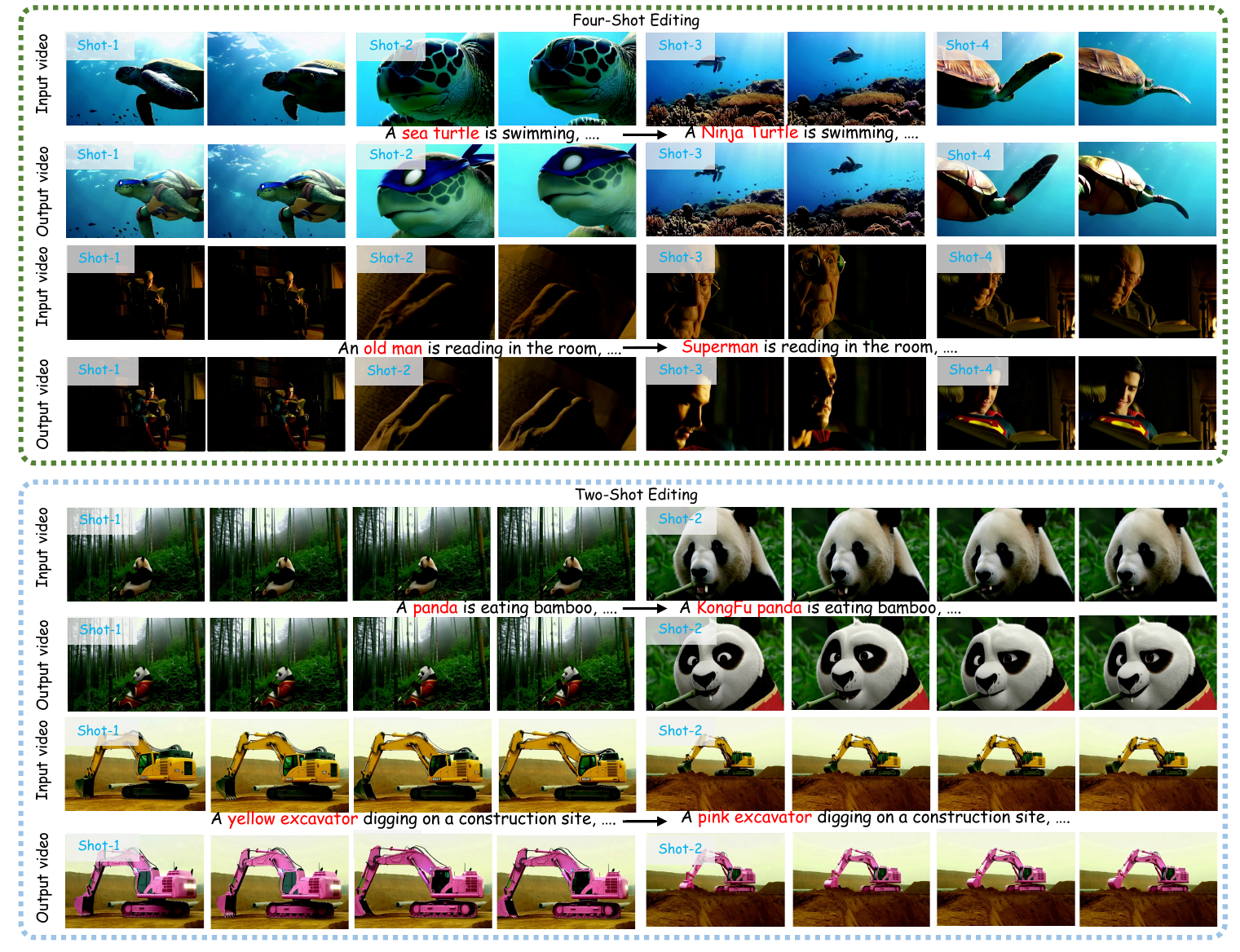}
  \caption{\textbf{Showcase of MSEditor.} Given a multi-shot input video, our method produces consistent edits across shots with diverse scene compositions, maintaining appearance, motion, and cross-shot consistency.}
  \label{fig:teaser}
\end{figure}

\section{Introduction}
\label{sec:intro}
Multi-shot video editing aims to achieve consistent and unified content modification across a sequence of discontinuous temporal segments (i.e., shots) assembled to convey a unified narrative. Unlike conventional single-shot video editing~\cite{tu2024motioneditor, liu2025sketchvideo, jiang2025vace, bian2025videopainter,ma2026group,wang2024taming,wang2024cove,ma2025magicstick}, which focuses on generating plausible modifications within a single continuous temporal flow, this task requires maintaining editing coherence in both appearance and structure across multiple shots. This capability is crucial for professional applications such as filmmaking, narrative storytelling, and commercial advertising, where visual content is inherently constructed from sequences of multiple diverse shots to create cinematic rhythm and convey complex storylines. Consequently, extending generative editing capabilities from single-shot to multi-shot scenarios is not merely an incremental step, but a crucial leap toward practical, professional-grade video creation.

While coherent multi-shot editing is highly desirable, ensuring semantic consistency across diverse shots remains a formidable challenge. These different shots encompass abrupt transitions in both spatial and temporal dimensions, including changes in viewpoints, camera scales (e.g., from a wide establishing shot to a close-up), and subject poses. Achieving high-level semantic consistency across these narratively linked but geometrically disjointed shots presents significant technical hurdles. Previous editing approaches typically operate on a single-shot basis~\cite{liu2024video, geyer2023tokenflow}. When applied to multi-shot sequences, they inherently struggle, suffering from severe identity drift and cumulative errors across temporal boundaries. Specifically, extending a single-shot video editing framework to multi-shot scenarios typically encounters two major bottlenecks:

%The primary limitation in extending a single-shot video editing framework to multi-shot tasks lies in maintaining subject consistency across discontinuous temporal boundaries. The previous video editing works always have two problems when they are applied to the multi-shot or long-sequence video editing scenarios: 

\textbf{(1) The lack of multi-shot video datasets.} A major bottleneck lies in the scarcity of large-scale, high-quality multi-shot video datasets. Constructing such datasets is extremely challenging, as it requires collecting sequences that preserve consistent subject identity across diverse scenes, lighting conditions, and transitions. Due to the absence of publicly available datasets that support robust multi-shot training, current diffusion-based video editing frameworks are often limited to single-shot settings and rely heavily on short-term temporal correlations. This restricts their generalization ability and hinders the development of multi-shot video editing.

\begin{figure}[t]
  \centering
  \includegraphics[width=0.9\linewidth]{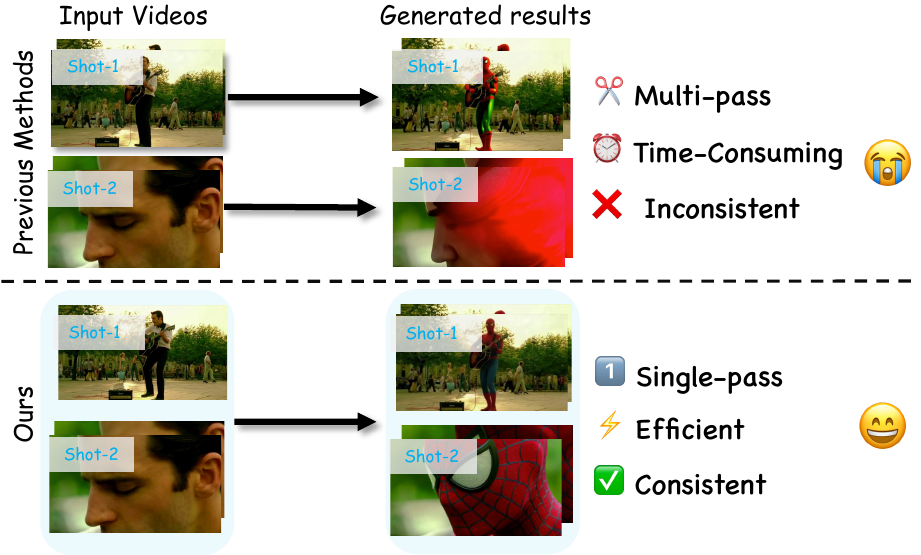}
  \caption{\textbf{Comparison with the previous single-shot editing framework}. (Top) Existing single-shot editing baselines partition multi-shot sequences into independent temporal chunks. This fragmented inference leads to cumulative errors and cross-shot inconsistency, while being computationally time-consuming due to multiple inference passes. (Bottom) Our MSEditor performs a single inference pass, achieving high-fidelity consistency and superior efficiency.}
  \label{fig:motivation_method}
\end{figure}
\textbf{(2) Cumulative error propagation and cross-shot inconsistency.} Within a localized temporal window (e.g., a short sequence containing two shots), previous editing frameworks~\cite{liu2024video, geyer2023tokenflow} often suffer from cumulative error propagation. Small editing artifacts or subtle identity drifts emerging in earlier frames tend to amplify recursively as the video progresses. When crossing shot boundaries, these accumulated deviations manifest as noticeable inconsistencies in the subject's appearance, style, and structural integrity. For longer videos that exceed the model's temporal context (e.g., 81 frames for Wan2.1~\cite{wan2025wan}), current paradigms~\cite{jiang2025vace,bai2025scaling} necessitate partitioning the sequence into independent temporal chunks for separate inference, as illustrated in Fig.~\ref{fig:motivation_method}. This fragmented processing fundamentally lacks cross-shot semantic awareness, causing severe subject identity drift between disjoint chunks. Furthermore, this approach is computationally inefficient, requiring multiple redundant inference passes that significantly increase the post-production overhead.

%In multi-shot sequences, small editing artifacts and identity drifts that appear in earlier shots tend to accumulate in subsequent ones. This progressive error propagation often results in noticeable inconsistencies in appearance, style, and expression of the same subject across shots. Consequently, the edited video may exhibit visual discontinuities that break the narrative flow and diminish the overall editing quality. As shown in Fig.~\ref{}, current single-shot editing paradigms~\cite{liu2024video, geyer2023tokenflow} necessitate partitioning a multi-shot sequence into independent temporal chunks for separate inference. This fragmented processing not only incurs significant computational redundancy through multiple inference passes but also fails to capture global semantic relationships. Consequently, minor editing artifacts in earlier shots often escalate into severe identity drifts and visual discontinuities in subsequent ones, breaking the narrative flow and degrading overall editing quality.
%Existing methods are primarily designed for single-shot videos~\cite{liu2024video,feng2024ccedit,geyer2023tokenflow} and lack explicit mechanisms to constrain cross-shot consistency. Therefore, they struggle to preserve the subject’s coherent appearance throughout long or complex video sequences.

% 不是收集multi shot 而是把现有的multi view看作是multi shot 然后用他们的数据来训练multi shot模型

To tackle these challenges, we propose \textbf{MSEditor}, a novel multi-shot video editing framework for coherent editing across diverse shots. To address the lack of multi-shot training data, we repurpose existing multi-view video datasets as a proxy for multi-shot supervision. Although they are originally designed for view synthesis, these datasets inherently contain sequences that capture the same subject under varying shots and scene conditions, providing valuable inter-shot correspondence for training. Specifically, we extract synchronized RGB sequences along with their associated depth maps, from which we compute accurate and temporally stable subject masks through projection-based consistency. This avoids the segmentation failures of conventional models such as SAM2~\cite{ravi2024sam}, which often misidentify background figures in complex scenes. Leveraging these structured multi-view signals, we introduce a Supervisory Adapter that injects cross-shot supervision into the backbone, enabling the model to learn subject relationships and maintain appearance consistency across shots.

To further reduce error accumulation and improve cross-shot coherence, we introduce a Cross-Shot Packing Strategy. Our method dynamically groups semantically related shots and processes them jointly within the self-attention window. This enables effective cross-shot feature interaction during training, facilitating subject identity preservation and ensuring consistent edits across diverse shots.
%To ensure the model does not overfit to privileged inputs, we randomly drop the adapter’s input with a fixed probability during training. This strategy encourages the backbone to internalize multi-shot-aware consistency while allowing fully geometry-free inference at test time.

%检测第一段和第二段的位置 做rope的对齐 提高一致性

In summary, our contributions are as follows:
\begin{itemize}
    \item We introduce \textbf{MSEditor}, the first framework specifically designed for the multi-shot video editing task, achieving robust identity preservation and coherent editing across different shots.
    \item We propose a Supervisory Adapter that leverages auxiliary priors to inject cross-shot information into the backbone, enabling the model to learn identity-consistent representations for multi-shot video editing.
    \item We design a Cross-Shot Packing Strategy that dynamically aligns features across shots and mitigates cumulative error propagation, ensuring long-range temporal and appearance consistency.
    \item Extensive experiments demonstrate the state-of-the-art performance of our method on multi-shot editing tasks, showing superior temporal stability and identity consistency over existing methods.
\end{itemize}

\section{Related Work}
\label{sec:related_work}
In this section, we review existing literature on single-shot video editing, multi-shot video generation, and long video generation.

\noindent\textbf{Single-Shot Video Editing.}
In recent years, video diffusion models have reshaped single-shot video editing~\cite{sun2024diffusion}. UNet-based approaches evolved from temporal mechanisms~\cite{chai2023stablevideo,couairon2023videdit,chen2025contextflow,liu2026tele} to training-free feature propagation via inter-frame correspondences~\cite{geyer2023tokenflow,ku2024anyv2v,song2024processpainter,wu2026vibe}. Other advances include attention manipulation for coherence~\cite{qi2023fatezero,liu2024video,ma2025followyourmotion,ma2025followfaster,article,wang2026liveedit}, adding spatiotemporal controls to image models~\cite{ceylan2023pix2video,wang2023zero}, and repurposing video models for editing~\cite{molad2023dreamix,yang2023rerender}. DiT-based methods exploit Diffusion Transformers~\cite{peebles2023scalable,ma2024followyouremoji,song2026vista} for scalability~\cite{zheng2024open,bai2025scaling}, with innovations in query-key manipulation~\cite{shen2025qk,wu2025freeswim,song2026streamingeffect,long2025follow,zheng2026forecast}, dataset construction~\cite{wu2025insvie}, unified in-context editing~\cite{ye2025unic,liang2026spongebob,jiang2025vace,Deng_2026_CVPR,deng2024compact}, and applications in autonomous driving~\cite{jiang2024dive}, inpainting~\cite{ma2025follow}, and controllable video editing~\cite{liu2025sketchvideo,ma2024followpose,ma2026fastvmt,gao2026pai,xu2026smrabooth,ma2025magicstick,liang2026cot,cao2026smart,feng2025dit4edit,ma2026livelightrealtimestreamingvideo}. However, these methods are designed for single shots and lack mechanisms for subject consistency across multiple shots with varying viewpoints.

\noindent\textbf{Multi-Shot Video Generation.}
Recently, multi-shot video generation has advanced via camera-controlled and camera-free approaches~\cite{ma2025controllable}. Camera-controlled methods model camera parameters for multi-view synchronization~\cite{bai2024syncammaster}, re-rendering~\cite{bai2025recammaster}, and pose control~\cite{he2024cameractrl}. Camera-free methods focus on narrative coherence: transition tokens for shot control~\cite{kara2025shotadapter}, Long Context Tuning to expand attention~\cite{guo2025long}, HoloCine for long-range consistency~\cite{meng2025holocine}, and SKALD for shot assembly~\cite{lu2025skald}. Specialized datasets include AnimeShooter~\cite{qiu2025animeshooter}, TalkCuts~\cite{chen2025talkcuts}, and Shot2story20k~\cite{han2023shot2story20k}. Despite the progress, these methods focus on generation from scratch rather than editing, and often require auxiliary geometric inputs, limiting practical use.

\noindent\textbf{Long Video Generation.}
Long video generation focuses on coherence over extended sequences~\cite{li2024survey,liu2025survey}. Early strategies used hierarchical latent diffusion~\cite{he2022latent} and parallel generation~\cite{yin2023nuwa}. Later works enabled flexible frame sampling~\cite{harvey2022flexible}, sampling correction~\cite{cheng2023consistent}, slice-based editing~\cite{cohen2024slicedit}, transition models~\cite{chen2023seine}, and noise rescheduling~\cite{qiu2023freenoise}. Recently, Mixture of Contexts (MoC)~\cite{cai2025mixture} uses sparse attention for near-linear scaling and minute-long coherence. However, these methods do not address the unique challenges of multi-shot editing, like cross-shot feature alignment and identity preservation across viewpoint changes.
\section{Method}

\begin{figure*}[t]
  \centering
  \includegraphics[width=0.92\linewidth]{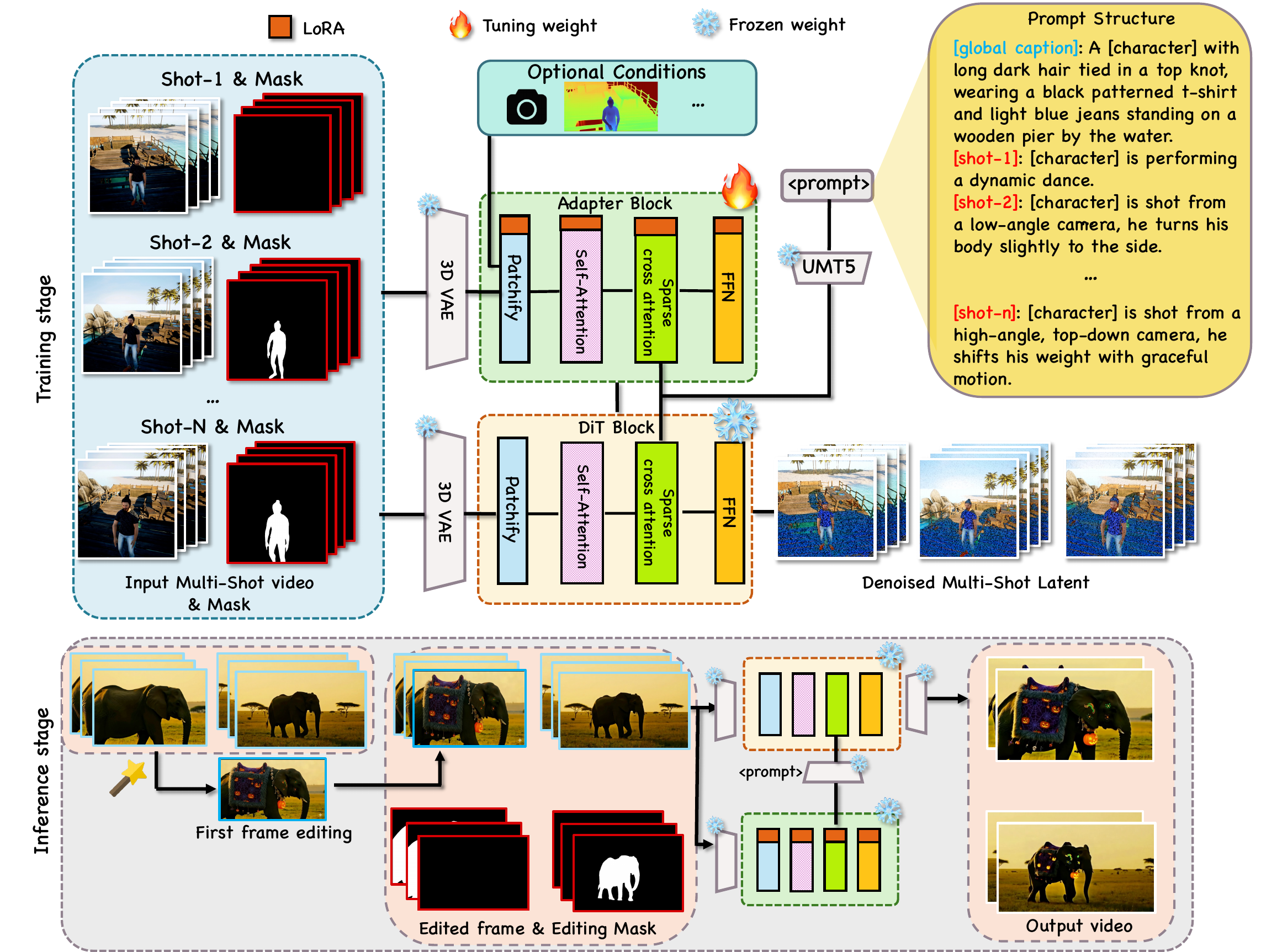}
  \caption{\textbf{Overview of our method.} \textbf{\textit{Top:}} Given an input multi-shot video and corresponding editing masks, our method first trains a Supervisory Adapter to inject multi-shot information into the diffusion backbone. We also introduce two key strategies: Cross-Shot Packing and Sparse Cross-Attention to keep the consistency during training. \textbf{\textit{Bottom:}} During inference, editing is performed on the first frame and produces coherent and identity-preserving multi-shot video output.
  }
  \label{fig:framework}
\end{figure*}
Our objective is to achieve temporally consistent and high-fidelity video editing across multiple discontinuous shots. 
% Sec.~\ref{sec:preliminary} reviews the foundational concepts of diffusion-based video generation models.
Sec.~\ref{sec:dataset} elaborates on our data construction and annotation approach. Sec.~\ref{sec:framework} presents our comprehensive framework for multi-shot video editing, which integrates a Supervisory Adapter for multi-shot information injection, a Cross-Shot Packing Strategy, and a Sparse Cross Attention Module for cross-shot consistency preservation.

\begin{figure*}[t]
    \centering
    \includegraphics[width=\linewidth]{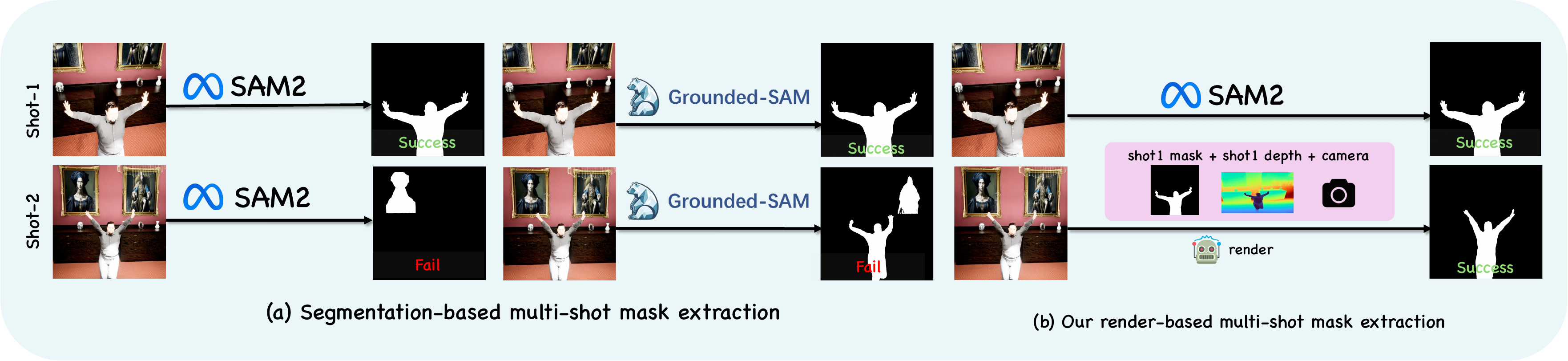}
    \caption{\textbf{Comparison of multi-shot mask extraction pipeline for training data construction.} (a) Using segmentation models (e.g., SAM2~\cite{ravi2024sam}, Grounded-SAM~\cite{ren2024grounded}) naively on the per-shot video often results in inconsistent results. The model may successfully segment the subject in one shot but fail in another due to a complex background. (b) Our proposed method bypasses this limitation by leveraging the multi-view data's camera and depth information. We use a complete mask from one shot and re-project it into another's view, rendering an accurate mask. This process ensures high-quality, cross-shot, consistent annotations.}
    \label{fig:segmentation}
\end{figure*}
\subsection{Data Annotation and Collection}
\label{sec:dataset}
The development of multi-shot video editing techniques faces a fundamental challenge: the scarcity of high-quality datasets capturing sequential, diverse shots of the same subject. Manual collection of such data is prohibitively labor-intensive, significantly constraining model training and generalization capabilities. 

To address this limitation, we repurpose existing multi-view video datasets~\cite{bai2024syncammaster,li2022neural} as an effective proxy. These datasets provide synchronized sequences of identical subjects from multiple viewpoints, containing rich inter-view correspondences that are instrumental for learning cross-shot consistency. 
However, adapting these datasets for multi-shot editing requires careful consideration. Unlike generation tasks, video editing demands precise subject-scene disentanglement to ensure accurate editing. As illustrated in Fig.~\ref{fig:segmentation}(a), direct application of off-the-shelf object segmentation models (e.g., SAM2~\cite{ravi2024sam}, Grounded-SAM~\cite{ren2024grounded}) often fails in complex scenarios where background elements contain distracting content such as paintings or incidental persons. 

To overcome this limitation, we develop a robust annotation method depicted in Fig.~\ref{fig:segmentation}(b). We employ double-reprojection techniques~\cite{yu2025trajectorycrafter} to generate reliable depth maps and view-consistent segmentation masks. These annotations provide explicit spatial and structural supervision during training, enabling precise subject isolation. To provide fine-grained semantic guidance, we implement an automated structured prompting pipeline using Qwen3-VL~\cite{Qwen3-VL}. For each multi-shot sequence, we generate a comprehensive global caption that describes the overarching subject identity and scene context. Additionally, we generate shot-specific descriptions for each viewpoint to capture unique camera angles and motion characteristics. This hierarchical annotation enables the model to learn both global semantic alignment and shot-level temporal dynamics.

The training dataset consists of 3,400 video sequences. For each video, there are synchronized RGB sequences from 10 different viewpoints $\{V^{(k)}\}_{k=1}^{10}$ paired with their corresponding segmentation masks $\{M^{(k)}\}_{k=1}^{10}$. The dataset also includes constructed depth maps $\{D^{(k)}\}_{k=1}^{10}$ and camera parameters $\{\text{cam}^{(k)}\}_{k=1}^{10}$. Each sequence is associated with a structured prompt: \begin{equation}
\text{Prompt} = \{\text{Caption}_{\text{Global}}, \text{Caption}_{\text{Shot-1}}, \text{Caption}_{\text{Shot-2}}, \dots, \text{Caption}_{\text{Shot-10}}\}.
\end{equation} This comprehensive annotation format enables the Supervisory Adapter to learn implicit, cross-shot consistent representations, ensuring both appearance consistency and high-fidelity editing across diverse shots.

%By combining RGB sequences with constructed depth and mask annotations, we assemble a multi-view editing dataset that serves as effective training data for our Supervisory Adapter. This dataset facilitates learning of implicit, cross-shot consistent representations essential for multi-shot video editing.

\subsection{Multi-Shot Video Editing Framework}
\label{sec:framework}
Our framework builds upon the pre-trained diffusion-based video generative backbone Wan2.1~\cite{wan2025wan}, which has demonstrated strong performance in video synthesis within single-shot scenarios. 
To enable coherent editing across different shots, we introduce \textbf{MSEditor}, a novel architecture designed to preserve subject identity throughout multi-shot sequences. The overall framework is illustrated in Fig.~\ref{fig:framework}.

% 待修改
\noindent\textbf{Supervisory Adapter for Multi-Shot Information Injection.} Although existing video editing frameworks exhibit strong editing capabilities in single-shot settings, they fail to maintain subject consistency across different shots due to the absence of multi-shot supervision. To enhance its multi-shot proficiency, we introduce a Supervisory Adapter that injects multi-shot features into the diffusion backbone during training. 

As shown in Fig.~\ref{fig:framework}, our adapter takes multi-view RGB videos and their corresponding segmentation masks as inputs. These multimodal signals guide the backbone to learn cross-shot consistent representations, enabling the model to preserve subject identity across discontinuous shots while maintaining high editing fidelity. Following the design principle of ControlNet~\cite{zhang2023adding}, all adapter layers are initialized to zero to ensure stable convergence and prevent over-conditioning.

To leverage powerful prior knowledge and enhance user-friendliness control, we follow the recent video editing works~\cite{ouyang2024i2vedit,liu2025sketch3dve} and adopt the first-frame editing paradigm that allows users to explicitly specify their editing intent on the initial frame. This reference-based approach significantly reduces the manual effort required for complex multi-shot sequences. Furthermore, to preserve the model's first-frame editing capability, we adopt a content editing strategy inspired by prior work~\cite{mou2024revideo}. Specifically, given a reference video 
\begin{equation}
\mathbf{V} = [\mathbf{I}_{0}, \dots, \mathbf{I}_{N-1}] \in \mathbb{R}^{N \times 3 \times H \times W},
\end{equation}
where $N$ denotes the number of frames, and $H$ and $W$ represent height and width respectively, we construct multi-shot video--mask pairs $\{(\mathbf{V}^{(k)}, \mathbf{M}^{(k)})\}_{k=1}^{K}$ corresponding to $K$ different shots.

During the training stage, the mask of the first frame $\mathbf{M}^{(0)}_0$ is set to zero for the first shot, establishing the first frame as the guidance during the video generation, while subsequent masks define the editable regions for propagation through the sequence. During inference, users modify the first frame, and the changes are propagated to the following frames. Overall, the training objective minimizes the MSE loss between predicted and ground-truth frames across all video--mask pairs:
\begin{equation}
\mathcal{L} = \frac{1}{K} \sum_{k=1}^{K} \left\| \mathbf{V}^{(k)}_{\text{pred}} - \mathbf{V}^{(k)}_{\text{gt}} \right\|_2^2.
\end{equation}
\begin{figure*}[t]
    \centering
    \includegraphics[width=\linewidth]{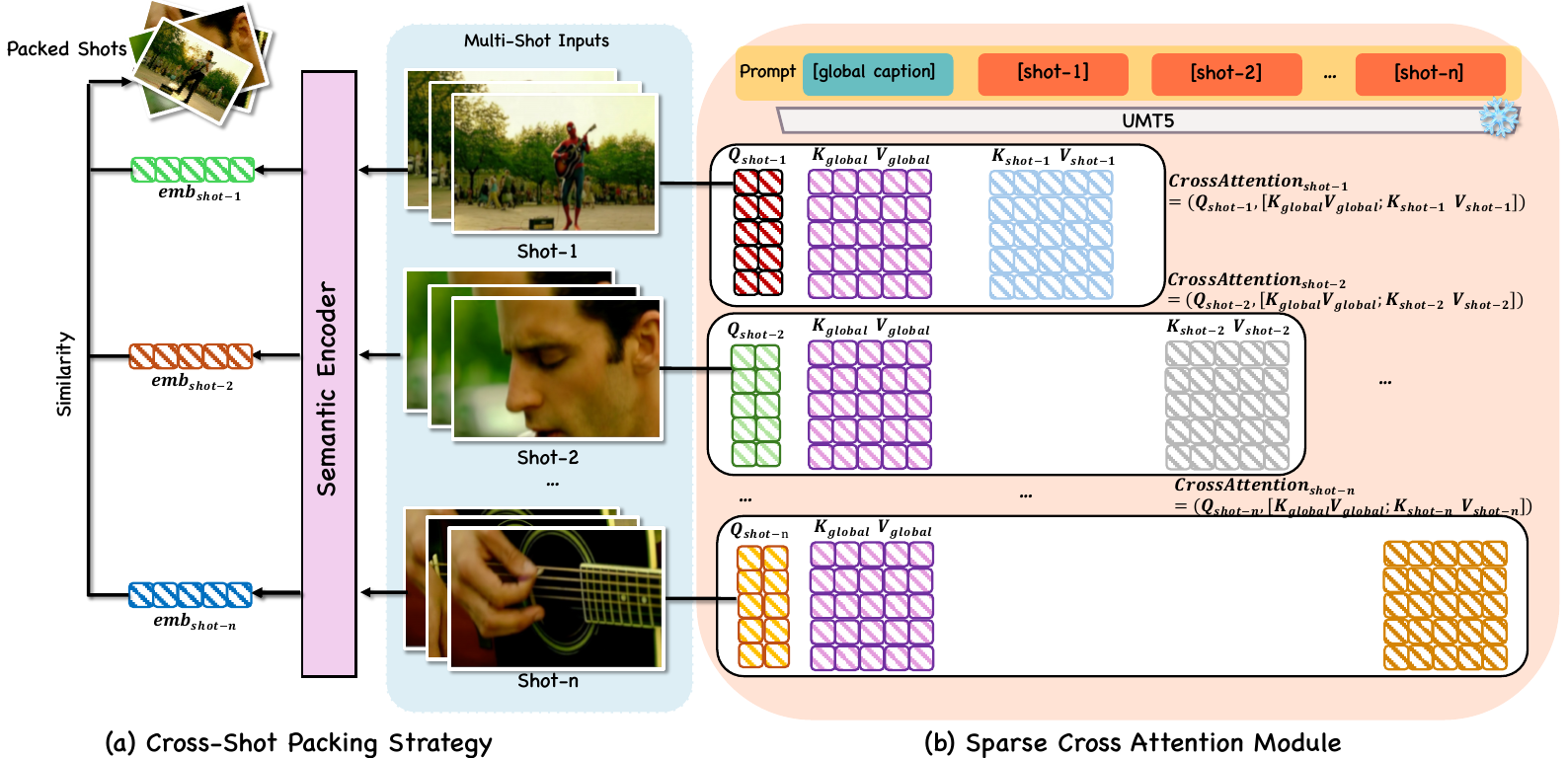}
    \caption{\textbf{Overview of the Cross-Shot Packing Strategy and Sparse Cross Attention Module.} (a) Given multi-shot inputs, the Cross-Shot Packing Strategy addresses the challenge of maintaining consistency in multi-shot video editing. (b) The Sparse Cross Attention Module integrates textual guidance. Visual queries from each shot (e.g., $Q_{\text{shot-1}}$) sparsely attend only to the concatenation of the shared global text embeddings ($K_{\text{global}}, V_{\text{global}}$) and their corresponding shot-specific embeddings ($K_{\text{shot-1}}, V_{\text{shot-1}}$).}
    \label{fig:attention}
\end{figure*}
This optimization enables the Supervisory Adapter to inject multi-shot priors into the backbone while ensuring the editing capability of the video foundation model, resulting in consistent multi-shot video editing.

\noindent\textbf{Cross-Shot Packing Strategy for Consistent Object Preservation.} While the Supervisory Adapter injects multi-shot information into our model, editing long videos with numerous shots remains challenging. In multi-shot scenarios, subjects may exhibit dramatic variations in scale, position, and appearance across shots. Minor inconsistencies in earlier shots can accumulate over time, leading to significant degradation of visual fidelity and structural coherence. However, in most training pipelines, multiple shots are typically concatenated along the batch dimension. Although this strategy improves training efficiency, it inherently isolates the temporal and semantic relations between different shots. Alternatively, concatenating all shots along the frame dimension would allow dense inter-shot interaction, but results in huge computational and memory costs.

To balance efficiency and consistency, we introduce a Cross-Shot Packing Strategy that reformulates self-attention computation by dynamically grouping semantically correlated shots into joint attention windows, as illustrated in Fig.~\ref{fig:attention}(a). Formally, given a multi-shot video $\{V^{(k)}\}_{k=1}^{K}$, we randomly select a shot $V^{\text{anc}} = \{f_{\text{anc},t}\}_{t=1}^{T_\text{anc}}$ as the anchor shot, which retains its full temporal resolution to serve as the primary training target. For all other candidate shots $V^{(j)}$, we compute their semantic similarity to the anchor using DINOv2~\cite{oquab2023dinov2} embeddings $e_{anc}$ and $e_j$. We then select the top-$M$ most correlated shots that exceed a threshold $\tau$ (e.g., 0.8). For each selected shot $V^{(j)}$, we extract a representative subset of $n$ frames (e.g., the initial frames) to serve as identity frames:\begin{equation}\hat{V}^{(j)} = \mathcal{S}(V^{(j)}, n), \quad \text{where } n < F_j.\end{equation}
The final packed sequence $\tilde{V}^{(\text{anc})}$ is then formulated by concatenating the full anchor shot with the distilled reference frames:\begin{equation}\tilde{V}^{\text{(anc)}} = \mathrm{Concat}\left(V^{\text{(anc)}}, \hat{V}^{(j_1)}, \dots, \hat{V}^{(j_M)}\right).\end{equation}

To ensure the training remains robust under various GPU memory constraints, we implement a dynamic thresholding strategy. If the total frame count of $\tilde{V}^{\text{(anc)}}$ exceeds the hardware's batch capacity, the system automatically increments $\tau$ (e.g., from 0.8 to 0.85) to filter out less relevant shots. This joint formulation enables the self-attention layers to perform feature interaction across semantically linked shots, facilitating a holistic understanding of subject identity across discontinuous temporal segments while maintaining computational tractability.

\noindent\textbf{Textual Conditioning via Sparse Cross Attention.} To integrate textual guidance, we adopt the Sparse Cross Attention mechanism from HoloCine~\cite{meng2025holocine} at both global and per-shot levels, as shown in Fig.~\ref{fig:attention}(b). Given token embeddings of the textual prompt, we separate them into a global caption and multiple per-shot descriptions:
\begin{equation}
\text{Prompt} = \{\text{Caption}_{\text{Global}}, \text{Caption}_{\text{Shot-1}}, \text{Caption}_{\text{Shot-2}}, \dots, \text{Caption}_{\text{Shot-K}}\}.
\end{equation}

For each shot $S_i$, we compute cross-attention between its visual queries $Q_i$ and the concatenated key--value pairs from global and corresponding per-shot textual embeddings:
\begin{equation}
    \mathrm{Attn}_{\text{cross}}(Q_i) = 
    \mathrm{Attention}\left(Q_i, [K_{\text{global}}, K_{i}], [V_{\text{global}}, V_{i}]\right).
\end{equation}

This formulation enables each shot to be guided by both global scene semantics and shot-specific textual context, thereby enhancing cross-shot consistency while preserving per-shot edit controllability.

\section{Experiments}
\begin{figure*}[t]
  \centering
  \includegraphics[width=\linewidth]{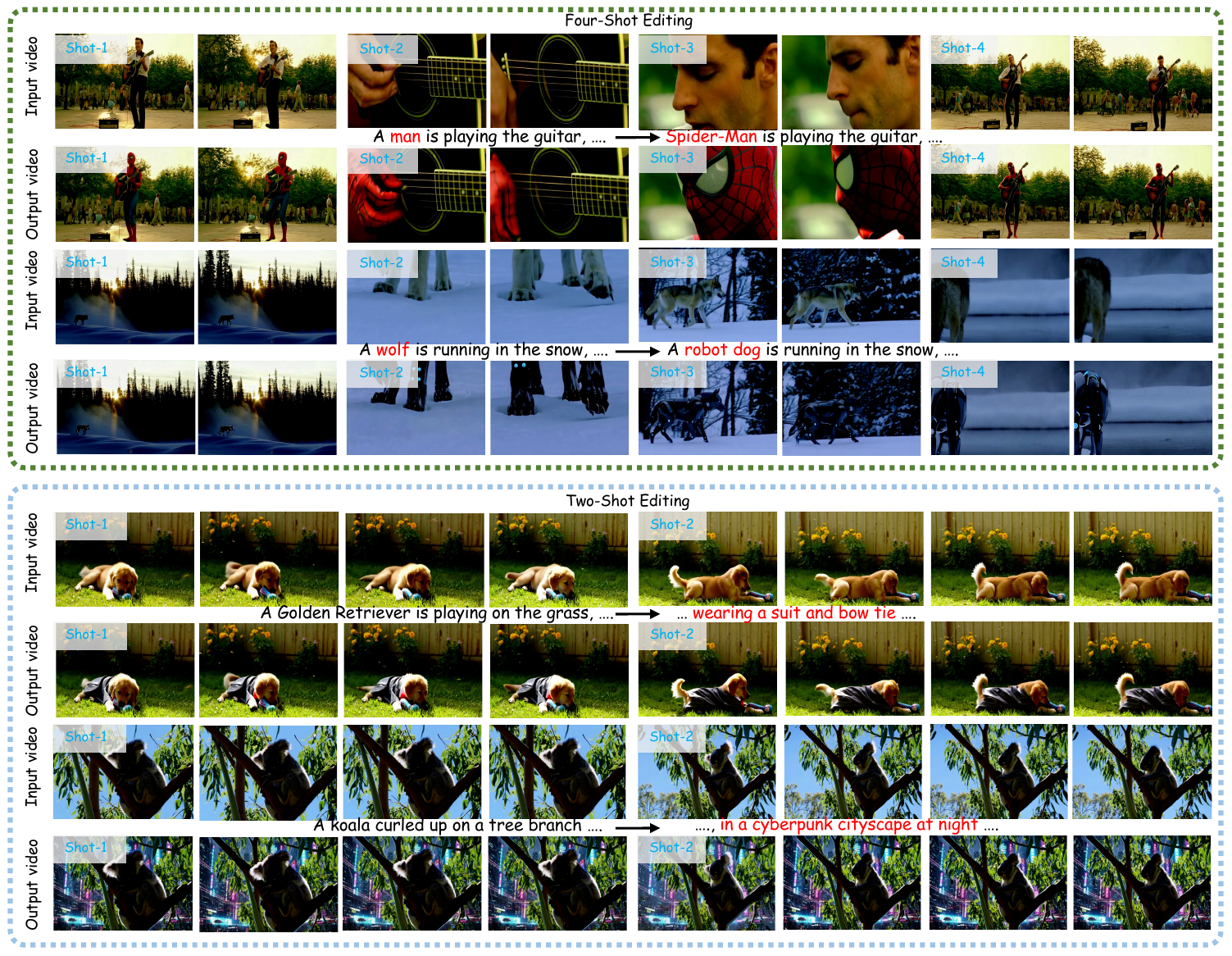}
  \caption{\textbf{Gallery of our proposed method, which displays four-shot and two-shot editing examples}. These results demonstrate the effectiveness of our model in applying diverse semantic modifications while maintaining high visual fidelity and cross-shot consistency. }
  \label{fig:gallery4}
  % \vspace{-0.6cm}
\end{figure*}

\begin{figure*}[t]
  \centering
  \includegraphics[width=\linewidth]{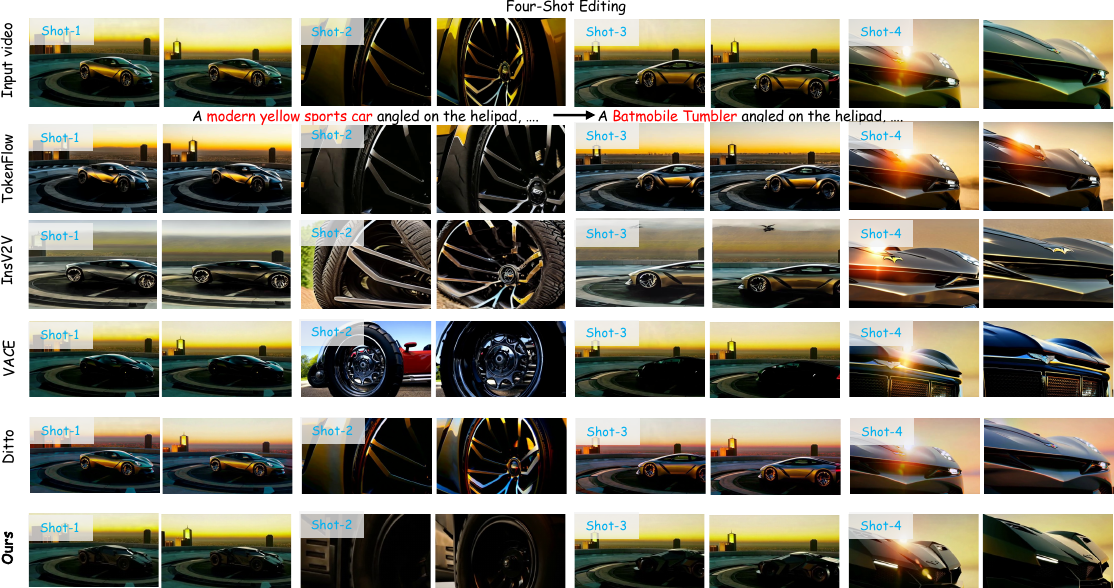}
  \caption{\textbf{Qualitative comparison results with the state-of-the-art methods.} We compare our multi-shot approach against leading single-shot video editing baselines on a four-shot editing task. The results show that our method successfully executes the complex subject replacement, demonstrating superior cross-shot temporal consistency. }
  \label{fig:compare}
\end{figure*}
\subsection{Experiment Setup}
\noindent\textbf{Datasets.} Our model is trained on the multi-shot dataset constructed through the methodology detailed in Sec.~\ref{sec:dataset}. For evaluation, we curate a diverse test dataset comprising 600 videos collected from the internet, featuring a balanced distribution of 35\% humans, 35\% animals, and 30\% objects. There is no overlap between our training and testing data, ensuring a rigorous assessment of the model's generalization capabilities. The evaluation tasks encompass style transfer, object replacement, and attribute editing, with prompts automatically generated via Qwen3-VL~\cite{Qwen3-VL}. More details of our test dataset are provided in the supplementary materials.

\noindent\textbf{Implementation Details.} We train our model based on Wan2.1-14B~\cite{wan2025wan}, a diffusion transformer-based video creation model. The optimization is performed using AdamW~\cite{loshchilov2017decoupled} with a weight decay of 0.01 and an initial learning rate of $1\times10^{-4}$. During the training process, we set the input resolution to $480 \times 832$ and a batch size of 8 on 8 NVIDIA A800 GPUs under PyTorch. During inference, we employ the flow-matching scheduler~\cite{lipman2022flow} with 50 sampling steps to generate high-quality and temporally coherent multi-shot videos. More evaluation metrics are provided in the supplementary material.

\begin{table*}[ht]
\centering
\caption{\textbf{Quantitative Results of Comparative Studies.} \textcolor{Red}{\textbf{Red}} and \textcolor{Blue}{\textbf{Blue}} denote the best and second best results, respectively.}

\resizebox{0.9\linewidth}{!}{\begin{tabular}{l|c|cc|cc|cc} 
\toprule
\multirow{3}{*}[0.8ex]{Method} & \multicolumn{1}{c|}{Video Quality} & \multicolumn{2}{c|}{Inter-shot Consistency}& \multicolumn{2}{c|}{Intra-shot Consistency} &\multicolumn{2}{c}{Semantic Consistency} \\
\cmidrule(lr){2-8} & Aesthetic Score $\uparrow$ & Subject $\uparrow$ & Background $\uparrow$ & Subject $\uparrow$ & Background $\uparrow$ &  Global $\uparrow$  & Shot $\uparrow$\\
\midrule

TokenFlow~\cite{geyer2023tokenflow} & 5.51 & 0.9181 & 0.9279 & 0.9029 & 0.9173 & 0.1302 & 0.1415 \\
InsV2V~\cite{cheng2023consistent} & 4.89 & 0.9013 & 0.9307 & 0.8962 & 0.8941 & 0.1433 & 0.1137\\
VACE~\cite{jiang2025vace} & \textcolor{Blue}{\textbf{5.91}} & 0.9237 & \textcolor{Blue}{\textbf{0.9372}} &  0.8805 & \textcolor{Blue}{\textbf{0.9365}} & \textcolor{Blue}{\textbf{0.1649}} & \textcolor{Blue}{\textbf{0.1738}}\\
Ditto~\cite{bai2025scaling} & 5.73 & \textcolor{Blue}{\textbf{0.9287}} & 0.9341 &  \textcolor{Blue}{\textbf{0.9125}} & 0.9331 & 0.1514 & 0.1569 \\

\cmidrule(lr){1-8}
\textbf{Ours}  & \textcolor{Red}{\textbf{6.05}} & \textcolor{Red}{\textbf{0.9370}} & \textcolor{Red}{\textbf{0.9447}}  & \textcolor{Red}{\textbf{0.9501}} & \textcolor{Red}{\textbf{0.9462}} & \textcolor{Red}{\textbf{0.1912}} & \textcolor{Red}{\textbf{0.1886}} \\
\bottomrule
\end{tabular}}

\label{tab:comparison}
\end{table*}

\begin{table*}[ht]
\centering
\caption{\textbf{Quantitative Results of Comparative Studies.} For fair comparison, we train recent methods VACE and VideoPainter on our dataset. \textcolor{Red}{\textbf{Red}} and \textcolor{Blue}{\textbf{Blue}} denote the best and second best results, respectively.}

\resizebox{0.9\linewidth}{!}{\begin{tabular}{l|c|cc|cc|cc} 
\toprule
\multirow{3}{*}[0.8ex]{Method} & \multicolumn{1}{c|}{Video Quality} & \multicolumn{2}{c|}{Inter-shot Consistency}& \multicolumn{2}{c|}{Intra-shot Consistency} &\multicolumn{2}{c}{Semantic Consistency} \\
\cmidrule(lr){2-8} & Aesthetic Score $\uparrow$ & Subject $\uparrow$ & Background $\uparrow$ & Subject $\uparrow$ & Background $\uparrow$ &  Global $\uparrow$  & Shot $\uparrow$\\

\cmidrule(lr){1-8}
VACE~\cite{jiang2025vace}& 5.93 & \textbf{\textcolor{Blue}{0.9301}} & 0.9395 & 0.9057 & \textbf{\textcolor{Blue}{0.9382}} & \textbf{\textcolor{Blue}{0.1728}} & \textbf{\textcolor{Blue}{0.1753}} \\

VideoPainter~\cite{bian2025videopainter}& \textbf{\textcolor{Blue}{6.01}}  & 0.9237 & \textbf{\textcolor{Blue}{0.9396}} & \textbf{\textcolor{Blue}{0.9061}} & 0.9377 & 0.1687 & 0.1732 \\

\midrule
\cmidrule(lr){1-8}
\textbf{Ours}  & \textcolor{Red}{\textbf{6.05}} & \textcolor{Red}{\textbf{0.9370}} & \textcolor{Red}{\textbf{0.9447}}  & \textcolor{Red}{\textbf{0.9501}} & \textcolor{Red}{\textbf{0.9462}} & \textcolor{Red}{\textbf{0.1912}} & \textcolor{Red}{\textbf{0.1886}} \\

\bottomrule
\end{tabular}}

\label{tab:comparison_train}
\end{table*}
\begin{figure}[ht]
    \centering
    \includegraphics[width=\linewidth]{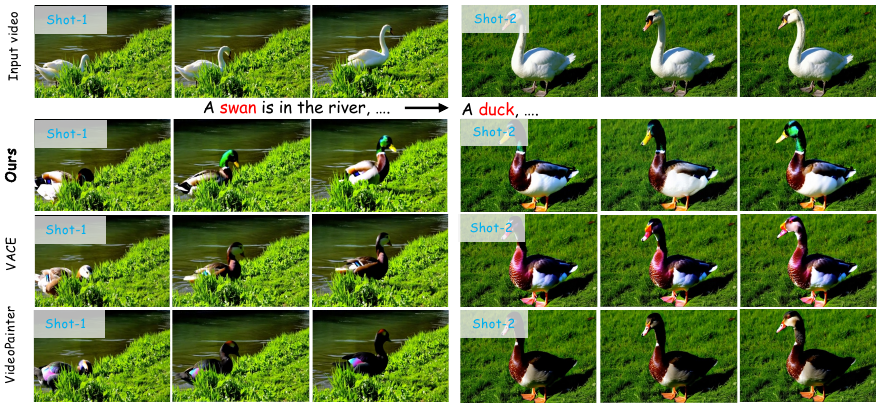}
    \caption{\textbf{More Qualitative Results after Training VACE and VideoPainter on Our Dataset.} For fair comparison, we train VACE and VideoPainter on our dataset, and compare the performance on the two-shot editing task. The results demonstrate that our proposed method has superior performance.} 
    \label{fig:com_trained}
\end{figure}

\subsection{Comparison with Baseline Methods}
\textbf{Qualitative Results.}
Due to the lack of existing methods for multi-shot video editing, we establish our baselines from the domain of single-shot video editing. To ensure a comprehensive comparison, we select methods spanning diverse architectures, namely UNet-based and DiT-based models. The chosen baselines include TokenFlow~\cite{geyer2023tokenflow}, InsV2V~\cite{cheng2023consistent}, Ditto~\cite{bai2025scaling}, and VACE~\cite{jiang2025vace}. Furthermore, to ensure a fair comparison, we retrained VACE~\cite{jiang2025vace} and VideoPainter~\cite{bian2025videopainter} on our curated multi-shot training dataset, enabling them to learn from identical data distributions. As shown in Fig.~\ref{fig:compare} and Fig.~\ref{fig:com_trained}, we display the qualitative results of these methods on four-shot and two-shot video editing tasks. Specifically, previous methods such as InsV2V, TokenFlow, and Ditto fail to perform the required edits while maintaining coherence. Even after being trained on our dataset, VACE and VideoPainter still struggle to maintain subject identity across discontinuous temporal boundaries, resulting in partial subject replacement or structural distortion. In contrast, our method demonstrates superior performance in editing fidelity, subject identity preservation, and temporal consistency. This indicates that the advantage of our approach stems not only from the training data but more importantly from our specialized architecture. Additional qualitative comparisons under more diverse settings are provided in the supplementary material. %As shown in Fig.~\ref{fig:gallery4}, we display the qualitative results of the existing single-shot video editing tasks on four-shot video editing tasks, and more qualitative results are in the supplementary material. Specifically, both InsV2V, TokenFlow, and Ditto fail to perform the edit, while VACE only achieves partial subject replacement that alters the subject's structure in a single shot. Compared with the previous editing methods, our method demonstrates superior results in editing fidelity, subject identity replacement, and temporal consistency when applied to multi-shot tasks.  

\noindent\textbf{Quantitative Results.}
For a quantitative study, we use the Aesthetic Score~\cite{schuhmann2022laion} to evaluate the quality of the edited video, and evaluate inter-shot consistency, intra-shot consistency, and semantic consistency following recent works~\cite{meng2025holocine}. As shown in Table~\ref{tab:comparison},  the quantitative results demonstrate that our proposed method outperforms the baseline across all metrics. To ensure a fair comparison, we retrained recent state-of-the-art methods, specifically VACE~\cite{jiang2025vace} and VideoPainter~\cite{bian2025videopainter}, using our own curated multi-view training dataset. As shown in Table~\ref{tab:comparison_train}, the quantitative results demonstrate that our proposed method outperforms all baselines across every metric. Notably, even after fine-tuning these baselines on the same data, our MSEditor still maintains a significant performance gap, which further validates the architectural superiority of our method. Furthermore, we provide a quantitative evaluation of our mask extraction accuracy and additional results for ``w/o mask setting'' and the shape-editing task in the supplementary material to further demonstrate our method's robustness. The results of our user study are also available in the supplementary material.%we collect 40 four-shot videos and 40 two-shot videos from the internet and perform assessments of the results obtained by our proposed method and the baseline. 

\begin{table*}[ht]
\centering
\caption{\textbf{Quantitative Results of the Ablation Study.} \textcolor{Red}{\textbf{Red}} and \textcolor{Blue}{\textbf{Blue}} denote the best and second best results, respectively.}
\resizebox{\linewidth}{!}{\begin{tabular}{l|c|cc|cc|cc} 
\toprule
\multirow{3}{*}[0.8ex]{Method} & \multicolumn{1}{c|}{Video Quality} & \multicolumn{2}{c|}{Inter-shot Consistency}& \multicolumn{2}{c|}{Intra-shot Consistency} &\multicolumn{2}{c}{Semantic Consistency} \\
\cmidrule(lr){2-8} & Aesthetic Score $\uparrow$ & Subject $\uparrow$ & Background $\uparrow$ & Subject $\uparrow$ & Background $\uparrow$ &  Global $\uparrow$  & Shot $\uparrow$\\

\cmidrule(lr){1-8}
w/o First Frame editing & \textcolor{Blue}{\textbf{6.01}}  & \textcolor{Blue}{\textbf{0.9321}} &\textcolor{Blue}{\textbf{0.9357}} & \textcolor{Blue}{\textbf{0.9336}} & \textcolor{Blue}{\textbf{0.9305}} &\textcolor{Blue}{\textbf{0.1752}}&\textcolor{Blue}{\textbf{0.1729}} \\
w/o Cross-Shot Packing strategy & 5.83 & 0.8919 & 0.9132 & 0.8937  & 0.9008 & 0.1738 & 0.1521 \\
w/o Sparse Cross attention & 5.70 & 0.8931 & 0.9256 & 0.8651 & 0.9126 & 0.1691 &0.1432 \\
\midrule
\textbf{Ours}  & \textcolor{Red}{\textbf{6.05}} & \textcolor{Red}{\textbf{0.9370}} & \textcolor{Red}{\textbf{0.9447}}  & \textcolor{Red}{\textbf{0.9501}} & \textcolor{Red}{\textbf{0.9462}} & \textcolor{Red}{\textbf{0.1912}} & \textcolor{Red}{\textbf{0.1886}} \\
\bottomrule

\end{tabular}}

\label{tab:ablation}
\end{table*}
% \begin{figure}[ht]
%     \centering
%     \includegraphics[width=\linewidth]{images/ablation.pdf}
%     \caption{\textbf{The qualitative Results of Ablation Study.} We ablate the proposed modules in two-shot editing tasks. Although the first shot is successfully edited, (1) Without the Cross-Shot Packing Strategy, the result lacks inter-shot consistency, leading to visual artifacts (e.g., a distorted face in the close-up shot). (2) Without the Sparse Cross Attention, the model fails in the second shot, indicating a failure to process the per-shot semantic guidance. (3) Without the Supervisory Adapter, the second shot merely re-colors the source object's structure instead of transforming it into the target, demonstrating a failure to maintain consistency.} 
%     \label{fig:ablation}
% \end{figure}
\subsection{Ablation Study}
\begin{figure}[t]
  \centering
  \includegraphics[width=0.9\linewidth]{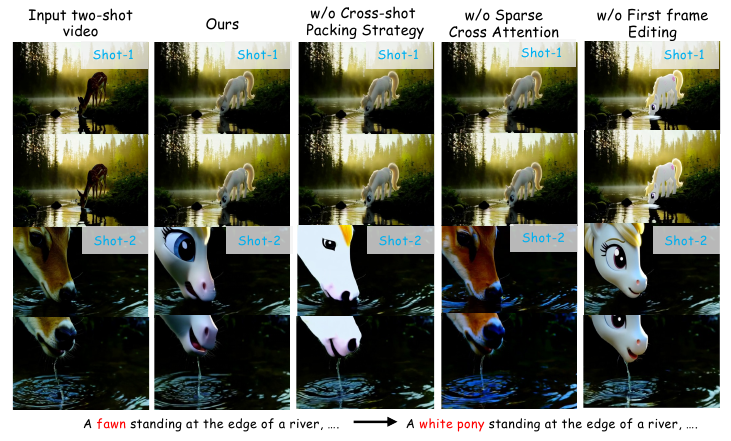}
  \caption{\textbf{Qualitative Results of the Ablation Study.} We ablate the proposed modules in two-shot editing tasks. Although the first shot is successfully edited, (1) Without the Cross-Shot Packing Strategy, the result lacks inter-shot consistency, leading to visual artifacts (e.g., a distorted face in the close-up shot). (2) Without the Sparse Cross Attention, the model fails in the second shot, indicating a failure to process the per-shot semantic guidance. (3) Without the First Frame Editing, the model successfully transforms the target but exhibits subtle deviations in fine-grained details, highlighting its role as a high-fidelity visual anchor rather than a prerequisite for structural consistency.}
  \label{fig:ablation}
\end{figure}

\noindent\textbf{Effectiveness of First Frame editing.} 
To investigate the contribution of the First Frame editing strategy, we conduct a series of ablation studies on it. In this setting, the model relies solely on the text prompt and the learned cross-shot priors to generate the edited content. The experimental settings are the same during the ablation. As shown in Fig.~\ref{fig:ablation} and Table~\ref{tab:ablation}, removing the first-frame anchor results in a slight decrease in consistency and aesthetic scores. However, this performance drop is relatively marginal compared to the impact of removing the other two strategies. This suggests that while first-frame editing serves as a high-fidelity visual anchor and significantly enhances user control, the core ability to maintain cross-shot consistency is fundamentally driven by our designed architecture. The results further demonstrate that our framework has learned robust, implicit semantic correspondences across disjoint shots.

\noindent\textbf{Effectiveness of Cross-Shot Packing strategy.} 
We further assess the effectiveness of the proposed Cross-Shot Packing strategy in Fig.~\ref{fig:ablation} and Table~\ref{tab:ablation}. Without the Cross-Shot Packing strategy, the model degenerates into the baseline training setting where multiple shots are concatenated along the batch dimension, and the edited video has the challenge of maintaining inter-shot consistency. 
\noindent\textbf{Effectiveness of Sparse Cross Attention.}
As shown in Table~\ref{tab:ablation}, removing Sparse Cross Attention leads to a clear degradation in inter-shot semantic consistency. In this setting, the model receives the same structured prompts, but we replace our designed module with a standard cross-attention mechanism. As shown in Table~\ref{tab:ablation}, this substitution leads to a marked degradation in inter-shot semantic consistency. Furthermore, in Fig.~\ref{fig:ablation}, we show the results when there is a lack of Sparse Cross Attention. This failure is primarily driven by severe semantic confusion. When processing the lengthy and structured concatenated prompts, a standard dense attention mechanism allows all video frames to attend globally to all text tokens. Consequently, the model struggles to disentangle the long context, often failing to accurately align specific shot-level textual instructions with their corresponding temporal segments.

% \begin{table*}[t]
% \centering
% \caption{\textbf{Ablation Study.} 
%  }

% \resizebox{0.9\linewidth}{!}{\begin{tabular}{l|c|cc|cc|cc} 
% \toprule
% \multirow{3}{*}[0.8ex]{Method} & \multicolumn{1}{c|}{Video Quality} & \multicolumn{2}{c|}{Inter-shot Consistency}& \multicolumn{2}{c|}{Intra-shot Consistency} &\multicolumn{2}{c}{Semantic Consistency} \\
% \cmidrule(lr){2-8} & Aesthetic Score $\uparrow$ & Subject $\uparrow$ & Background $\uparrow$ & Subject $\uparrow$ & Background $\uparrow$ &  Global $\uparrow$  & Shot $\uparrow$\\
% \midrule
% w/o Supervisory Adapter & 5.94  & 0.9001 & 0.9176 & 0.8821 & 0.9056 &0.1752&0.1558 \\
% w/o Cross-shot attention & 5.83 & 0.8923 & 0.9135 & 0.8942  & 0.9017&0.1741&0.1532 \\
% w/o Sparse Cross attention & 5.72 & 0.8935 & 0.9268 & 0.8657 & 0.9135&0.1698&0.1439 \\
% \midrule

% \textbf{Ours}  & \textcolor{Red}{\textbf{6.07}} & \textcolor{Red}{\textbf{0.9374}} & \textcolor{Red}{\textbf{0.9451}}  & \textcolor{Red}{\textbf{0.9513}} & \textcolor{Red}{\textbf{0.9479}} & \textcolor{Red}{\textbf{0.1924}} & \textcolor{Red}{\textbf{0.1892}} \\
% \bottomrule
% \end{tabular}

% \label{tab:ablation}

% }
% \end{table*}

\section{Conclusion}
We have proposed \textbf{MSEditor}, the first multi-shot video editing framework. Unlike prior single-shot methods that suffer from subject drift and temporal inconsistency, our approach leverages structured multi-shot supervision during the training stage to endow the diffusion model with cross-shot awareness. Specifically, our Supervisory Adapter injects multi-shot information into the backbone, facilitating the structural coherence of the model. Additionally, we design the Cross-Shot Packing strategy to keep the consistency between different shots. Extensive experiments demonstrate that our approach substantially improves cross-shot alignment and subject fidelity compared to previous diffusion-based editing frameworks. We believe our framework opens promising avenues for future research in real-world, multi-shot storytelling applications.

\section{Acknowledgments}
This research was supported by the National Natural Science Foundation of China (No. 62502410), the Guangdong Basic and Applied Basic Research Foundation (No. 2026A1515011138) and the Research Grants Council of HKSAR under grant number AoE/E-601/24-N.
% \clearpage

\bibliographystyle{unsrtnat}
\bibliography{main}

\end{document}